\documentclass[10pt,twocolumn,letterpaper]{article}

\usepackage{wacv}              

\usepackage{graphicx}
\usepackage{amsmath}
\usepackage{amssymb}
\usepackage{booktabs}
\usepackage{times}
\usepackage{epsfig}
\usepackage{graphicx}
\usepackage{amsmath}
\usepackage{amssymb}
\usepackage{booktabs}
\usepackage{xcolor}
\usepackage{threeparttable}
\usepackage{enumitem}
\usepackage{multirow}
\usepackage{siunitx}
\usepackage{math}
\usepackage{pifont}
\usepackage{makecell}
\usepackage{arydshln}

\usepackage{caption}
\usepackage{subcaption}

\setlist{nolistsep}

\def\eg{\emph{e.g.}} 
\def\etal{\emph{et al.}} 
\def\ie{\emph{i.e.}}

\def\method{S2P}
\def\kd{\emph{KD}}

\usepackage[pagebackref,breaklinks,colorlinks]{hyperref}

\usepackage[capitalize]{cleveref}
\crefname{section}{Sec.}{Secs.}
\Crefname{section}{Section}{Sections}
\Crefname{table}{Table}{Tables}
\crefname{table}{Tab.}{Tabs.}

\def\wacvPaperID{95} 
\def\confName{WACV}
\def\confYear{2024}

\begin{document}

\title{Source-Guided Similarity Preservation for Online Person Re-Identification}

\author{Hamza Rami$^{1,2}$,
        Jhony H. Giraldo$^{1}$,
        Nicolas Winckler$^{2}$,
        Stéphane Lathuilière$^{1}$\\
$^{1}$LTCI, Télécom Paris, Institut Polytechnique de Paris.\\
$^{2}$Atos.\\
\tt\small \{hamza.rami, jhony.giraldo, stephane.lathuiliere\}@telecom-paris.fr, \tt\small nicolas.winckler@atos.net     
}
\maketitle

\begin{abstract}
Online Unsupervised Domain Adaptation (OUDA) for person Re-Identification (Re-ID) is the task of continuously adapting a model trained on a well-annotated source-domain dataset to a target domain observed as a data stream.
In OUDA, person Re-ID models face two main challenges: catastrophic forgetting and domain shift.
In this work, we propose a new Source-guided Similarity Preservation (\method) framework to alleviate these two problems.
Our framework is based on the extraction of a support set composed of source images that maximizes the similarity with the target data.
This support set is used to identify feature similarities that must be preserved during the learning process.
\method~can incorporate multiple existing UDA methods to mitigate catastrophic forgetting.
Our experiments show that \method~outperforms previous state-of-the-art methods on multiple real-to-real and synthetic-to-real challenging OUDA benchmarks.
\end{abstract}

\section{Introduction}
\label{sec:intro}

Person Re-Identification (\textit{Re-ID}) is the task of recognizing a person of interest (\ie, query) across a set of images taken by non-overlapping cameras (gallery) \cite{Ye2022DeepLF}.
Person Re-ID has attracted a lot of interest because of the rising need for public safety and intelligent surveillance systems.
Recently, the accuracy of Re-ID models has significantly improved when using supervised deep learning \cite{CCS}.
However, the performance of these approaches drastically decreases when they are deployed in data that visually differ from the training dataset \cite{BoT}.
Since collecting data for every new environment is not practical, previous studies have introduced Unsupervised Domain Adaptation (\emph{UDA}) for person Re-ID \cite{MMT, SpCL, PTGAN}.

\begin{figure}
    \centering
    \includegraphics[width=\columnwidth]{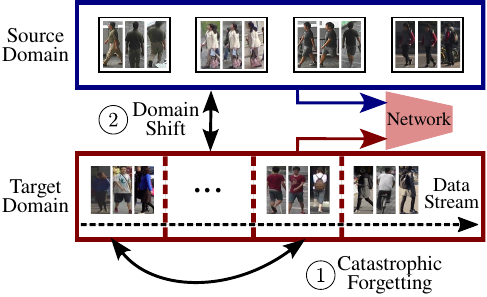}
    \caption{In OUDA for person Re-ID, the images of the target domain are available as a stream of data, and past images cannot be stored. Two main challenges should be addressed: 1) catastrophic forgetting and 2) domain shift.}
    \label{fig:catastrophic_forgetting_domain_shift}
\end{figure}

UDA methods combine a well-annotated dataset (\textit{source domain}) and an unlabeled dataset corresponding to the \textit{target domain}, aiming to train a model that can perform well in the new environment.
Despite progress in recent years~\cite{MMT,SpCL}, UDA for person Re-ID suffers from three main issues that prevent its practical use.
First, when collecting the target data required to adapt the model, images are generally gathered as a stream that continually sends photos from various cameras/locations. Consequently, collecting a large target dataset may take time and delay deployment.
In addition, in UDA, the model is frozen after deployment and does not benefit from the new data, which are continuously captured.
Finally, numerous countries have adopted privacy regulations that forbid technology providers to store images of individuals.
Thus, collecting a large target dataset is not possible.

Since deploying algorithms that conform with policies of data privacy protection has become a legal obligation in a growing number of countries, the Online Unsupervised Domain Adaptation for person Re-Identification (\textit{OUDA-Rid}) setting was introduced in \cite{cvprw} to address the limitations of traditional UDA techniques. In the OUDA-Rid framework, we operate under the assumption that we have access to annotated source data as well as unlabeled target data. However, in contrast to traditional UDA settings, the target dataset is treated as an online stream of data, aligning with the constraint that camera-captured images cannot be stored. In addition to complying with privacy-protection regulations, this setting also enables the person Re-ID model to be continuously updated as new target data becomes available, thereby improving its adaptability to changes in the target domain.





{The performance of existing UDA methods for OUDA-Rid shows significant drops in performance regarding the offline setting \cite{cvprw}.}
This drop can be explained by the two main difficulties of OUDA-Rid illustrated in Fig.~\ref{fig:catastrophic_forgetting_domain_shift}: catastrophic forgetting and domain shift.
Catastrophic forgetting appears when the model only observes a few target identities, and consequently, the model tends to forget previously learned identities.
Domain shift is a change in the data distribution between the source and target domains. Addressing the domain shift is especially challenging in the online setting since, at every training step, we observe only a small and possibly biased subset of the target domain.


In this work, we consider that these two difficulties must be addressed jointly since mitigating catastrophic forgetting can lead to target representations that better capture the full target distribution, and consequently facilitate source-target distribution alignment.
We introduce a unified Source-guided Similarity Preservation (\textit{\method}) framework for OUDA-Rid that addresses these two challenges jointly.
We take inspiration from \emph{replay-based} strategies \cite{continuously_shifting_domains,ContrastiveRehearsal} to introduce a Knowledge Distillation (\kd) mechanism. By transferring the knowledge acquired with a teacher model to a student model, the KD \cite{Hinton} method enables the learning of more robust and generalizable features.
However, unlike existing replay-based approaches, we do not store any target image to conform to the \emph{privacy protection} requirement.
To this end, we extract a support set composed of source images that are similar to the previously seen images of the target.
This support is thus used to regularize the learning process and alleviate catastrophic forgetting.
Our framework combines both explicit source-target distribution alignment and pseudo-labeling to address domain shift.
\method~can easily integrate almost any existing UDA approaches \cite{MMT, SpCL} and readily outperforms all state-of-the-art methods for OUDA-Rid in several challenging conditions in real-to-real and synthetic-to-real tasks.
Our main contributions can be summarized as follows:
\begin{itemize}[leftmargin=*]
    \item We introduce a novel \method~algorithm that uses source-guided similarity preservation to jointly alleviate the \textit{catastrophic forgetting} and \textit{domain shift} while respecting the \emph{privacy protection} requirements. 
    \item \method~can easily incorporate almost any existing UDA approach.
    In particular, we present the integration of the \emph{MMT} \cite{MMT}, \emph{SpCL} \cite{SpCL} and \emph{IDM} \cite{IDM} methods into our framework, which achieve remarkable results in the UDA setting.
    \item We perform extensive experiments\footnote{Code available:  \href{https://github.com/ramiMMhamza/S2P}{https://github.com/ramiMMhamza/S2P}} in real-to-real and synthetic-to-real OUDA tasks with four datasets.
    \method~readily improves previous state-of-the-art UDA methods for OUDA-Rid.
    A set of ablation studies validate each component of our algorithm.
\end{itemize}
\section{Related Work}
\label{sec:related}



\noindent \textbf{UDA for person Re-ID.} Existing methods can be divided into \textit{domain translation-based} and \textit{pseudo-labeling}.

\emph{Domain translation-based} methods~\cite{ATN,SDA,IGC} modify the source domain images to resemble the appearance of the target set with style transfer approaches~\cite{CycleGan}.
Recent research focuses on enhancing translation by preserving self-similarity~\cite{SPGAN} or performing camera-specific translation~\cite{HHL}.

\emph{Pseudo-labeling} methods employ an iterative process alternating between clustering and fine-tuning \cite{theory&practice, DGM, BottomUp, CPS, delorme2021canu}.
Fan \etal~\cite{SB} proposed a simple and effective baseline where the Re-ID model is fine-tuned using cluster indices as labels.
Several studies have expanded on this framework, such as self-similarity grouping \cite{SSG}, Mutual-Mean Teaching (\textit{MMT}) \cite{MMT}, and Self-paced Contrastive Learning (\textit{SpCL}) \cite{SpCL}.
MMT adopts a teacher-student framework where two student networks are jointly trained using pseudo-labels generated by themselves and soft pseudo-labels generated by their mean-teacher networks.
On the other hand, SpCL takes a different approach by gradually constructing more reliable clusters to refine a hybrid memory containing both source and target images. 
More recently, the use of an Intermediate Domain Module (IDM) \cite{IDM} has also been explored as means to bridge the gap between source and target domains.

{We adopt the pseudo-labeling framework as it has outperformed previous techniques in almost all datasets \cite{MMT,SpCL} and avoids the computational overhead of transfer-based methods.
Our \method~overall framework can incorporate existing pseudo-labeling methods toward better performance in the OUDA setting.}

\noindent\textbf{Lifelong learning for Re-ID.} Lifelong learning, also called Continual Learning (\textit{CL}) or incremental learning \cite{Online1, Online2, Online3}, is a field that aims at developing adaptive agents, like the way humans learn throughout their lifetime.
The main problem of CL is \textit{catastrophic forgetting}, meaning that the model tends to forget the previously acquired knowledge.
Recently, several methods have been developed to solve this issue in typical vision tasks \cite{onlineod,onlineseg,lifelonggan}.
We can categorize existing lifelong learning approaches into three main categories: 1) teacher-student \cite{LWF2, lifelongts}, 2) regularization \cite{lifelong_learning}, and 3) replay methods \cite{MRGAN}.

\begin{figure*}
    \centering
    \begin{subfigure}[t]{0.70\textwidth}
        \centering
        \includegraphics[height=1.9in]{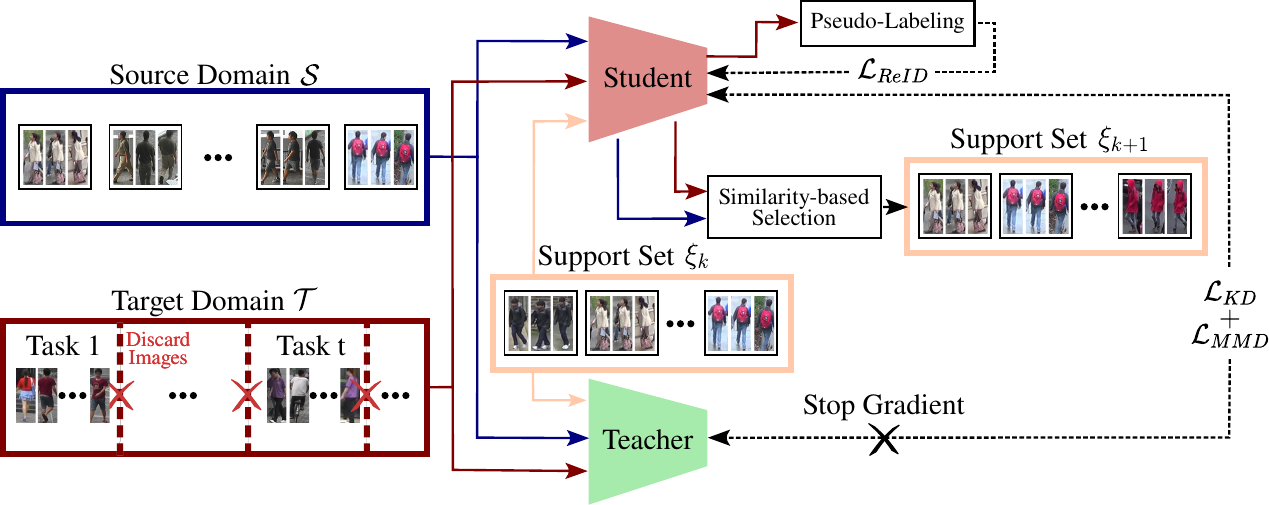}
        \caption{Overall Pipeline of \method.}
    \end{subfigure}%
    \hfill
    \begin{subfigure}[t]{0.30\textwidth}
        \centering
        \includegraphics[height=1.9in]{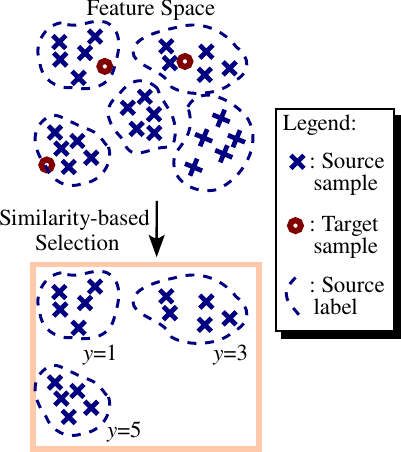}
        \caption{Construction of the support set via similarly-based selection.}
    \end{subfigure}
    \caption{The pipeline of \method. a) \method~incorporates knowledge distillation $\mathcal{L}_{KD}$, discrepancy $\mathcal{L}_{MMD}$ loss functions, and a teacher model to mitigate the catastrophic forgetting and domain-shift problems. b) Our algorithm employs a similarity-based selection to construct the support set $\xi_k$ from the source domain that maximizes the similarity with the target images.}
        \label{fig:framework}
\end{figure*}

Few studies have tackled the problem of lifelong learning in person Re-ID.
For instance, Pu \etal~\cite{AKA} proposed an Adaptive Knowledge Accumulation (\textit{AKA}) framework, which is fully supervised.
AKA addresses the domain-incremental setting where each task corresponds to a different target domain.
Huang \etal~\cite{lifelong_uda_reid} also adopt an incremental scenario, although storing images from the previous task is permitted.

While previous methods in continual learning for person Re-ID, such as \cite{AKA,lifelong_uda_reid}, have adopted a less restrictive setting that allows keeping images from previous tasks, we follow the more challenging and privacy-preserving OUDA-Rid setting proposed in \cite{cvprw}.
To address domain shift and catastrophic forgetting in OUDA-Rid, we introduce two key technical contributions: a source-guided knowledge distillation strategy and an explicit domain alignment.
Gong \etal~\cite{gong2013connecting} introduced a technique based on landmarks that is similar to our support set selection.
However, these landmarks were proposed to solve the domain gap in the context of UDA with classical machine learning techniques, while we have to also consider the catastrophic forgetting problem in OUDA-Rid using end-to-end deep learning models.

\section{Source-Guided Similarity Preservation}
\label{sec:method}


\noindent\textbf{OUDA-Rid problem definition.} In OUDA-Rid, we assume having access to a well-annotated source domain dataset $\mathcal{S}=\{(\xvect_{i}^{S},\yvect_{i}^{S})\}_{i=1}^{Ns}$, and an unlabeled target domain dataset $\mathcal{T} = \{ \xvect_i^t \}_{i=1}^{Nt}$. Here, both domain images are not necessarily drawn from the same distribution. We consider that we have access to the target domain dataset in the form of an ongoing stream of data.
Following the common batch-based approximation of the online learning setting~\cite{fini2020online}, we consider that we observe a sequence of $N_T$ tasks $\{{\mathcal{T}^1} \cup {\mathcal{T}^2} \cup ... \cup {\mathcal{T}^{N_T}}\}$. 
Each task $\mathcal{T}^k,1\leq k\leq N_T$ is a set of images captured by several cameras and depicting an unknown number of identities.
To align with practical scenarios, we consider that each identity can be observed by different cameras simultaneously. However, it is unlikely for an identity to appear at widely separated time intervals (\eg~different days). Therefore we can assume that identities do not overlap across different tasks, although this assumption is not strictly required in our approach.

In the rest of this section, we present our \method~framework to alleviate the two major challenges of the OUDA setting: catastrophic forgetting and domain shift.
First, our framework integrates a teacher model that distills previously acquired knowledge.
The KD strategy of \method~is based on feature space similarity preservation and only requires images from the source domain, hence respecting the \emph{privacy protection} norms.
Second, we minimize the discrepancy between the source domain and the target domain to reduce the domain shift and further enhance the stability of the \method.

\subsection{Overview of the Approach}

Fig. \ref{fig:framework} shows the pipeline of our \method~framework.
In every task {of the OUDA-Rid problem}, the target labels are not available and we assume that the identities are different even if our \method~does not strictly require this assumption.
Furthermore, we construct a \textit{support set} that plays the role of a memory bank for \textit{source-guided knowledge distillation}.
We could keep a few samples from previous tasks if there were no privacy constraints.
However, in \method~the support set only includes images from the source domain.
We choose those images based on their similarities to previously seen images, ensuring a good approximation of the previously learned feature spaces during continual learning.

In this work, we follow an overall training scheme that was adopted by multiple UDA methods for Re-ID~\cite{MMT,SpCL}. 
More concretely, we use a student model that consists of a feature extractor $\mathcal{F}(\cdot)$. 
First, the student model is pre-trained on source data $\mathcal{S}$, and then fine-tuned on the unlabeled target data $\mathcal{T}$ with three different loss functions:
\begin{itemize}[leftmargin=*]
    \item $\mathcal{L}_{KD}$: the knowledge distillation loss in the feature similarity space is proposed to preserve the previously acquired knowledge.
    To this end, a \textit{similarity-based selection} strategy is applied to the source domain to construct the support set, and a teacher model $\bar{\mathcal{F}}(\cdot)$ is added to the main pipeline (Sec. \ref{sec:kd}).
    \item $\mathcal{L}_{MMD}$: the Maximum Mean Discrepancy (\textit{MMD}) loss is minimized to reduce explicitly the domain shift.
    In other words, we want to construct a feature space that is domain invariant and can regroup features from both the source and the target domains (Sec. \ref{sec:MMD}).
    \item $\mathcal{L}_{ReID}$: this loss corresponds to the loss of the UDA method that is integrated into our framework.
    This loss is jointly minimized on the source domain $\mathcal{S}$ and the target domain images $\mathcal{T}$ together with their pseudo-labels.
    The pseudo labels are estimated by a clustering algorithm assigning each image to a cluster label (Sec. \ref{sec:IoM}). 
\end{itemize}

\subsection{Source-Guided Knowledge Distillation} 
\label{sec:kd}

When learning a new task $\mathcal{T}^k$, the model must be updated to better discriminate the appearance of the new individuals.
However, the model should also preserve the knowledge acquired on previous tasks $\mathcal{T}^i~\forall~1\le\!i\le\!k-1$.
Therefore, we employ a teacher model that progressively distills the knowledge to the student model. 
Distillation is performed in the feature space over a set of source-based support images.
Since target images cannot be stored, we propose to use images from the source domain as the support set.
More precisely, we select images that are similar to the images from the target domain seen in previous tasks.
This solution encourages the student model to project the images into a common feature space, resulting in more discriminant and task-invariant representations.

\noindent\textbf{Support set collection.}
Fig. \ref{fig:framework} (b) depicts the construction of the support set in \method.
We construct the support set based on the cosine similarity in the feature space between the current target images and the source domain images. 
For each image $\xvect^t$ in the target task $\mathcal{T}^k$, we identify the image $\mathcal{\xi}_x(\xvect^t)$ and its corresponding identity label $\mathcal{\xi}_y(\xvect^t)$ from the source domain that maximizes the cosine similarity in the feature space:
\begin{equation}
     \left(\mathcal{\xi}_x(\xvect^t),\mathcal{\xi}_y(\xvect^t)\right) = \argmax_{(\xvect^s,\yvect) \in \mathcal{S}}\frac{\mathcal{F}(\xvect^s)\cdot\mathcal{F}(\xvect^t)}{\lVert \mathcal{F}(\xvect^s)\rVert\lVert \mathcal{F}(\xvect^t)\rVert}.
\end{equation}
Then, we add to the support set all the images from the source that correspond to the selected identity $\mathcal{\xi}_y(\xvect_t$):
\begin{equation}
\label{eq:support-set}
    \mathcal{\xi}_k = \bigcup_{\xvect^t \in \mathcal{T}^k} \{(\xvect^s,\yvect) \in  \mathcal{S}, \yvect = \mathcal{\xi}_y(\xvect^t)\}.
\end{equation}
While learning a new task $\mathcal{T}^{k+1}$, $\mathcal{\xi}_k$ is used as a memory that best approximates previously seen images.


\noindent\textbf{Teacher-student framework.} 
As a teacher, we need a model that has accumulated knowledge from previous tasks and can effectively guide the student's learning on a new task.
We use the Exponential Moving Average (\textit{EMA}) parameters update \cite{Mean_Teachers, TempEns} of the current model. 
At every iteration $i$, the parameters $\bar{\thetavect}_{i}$ of the teacher model are given by:
\begin{equation}
    \label{eq:teacher}
   \bar{\thetavect}_{i} = \alpha \bar{\thetavect}_{i-1} + (1-\alpha) \thetavect, 
\end{equation}
where $\thetavect$ denotes the current parameters of the student model and $\alpha \in [0,1)$ is the weighting factor. 
At the first iteration of our framework, $\bar{\thetavect}_{0}$ is initialized using a model pre-trained on the source dataset.
Once the adaptation process is performed on a specific task, only the teacher is used for inference on the test set.
\def\fbvect{\boldsymbol{\bar{f}}}
\def\fbmat{\mbox{\bf f}}

\noindent\textbf{KD loss.}
Knowledge distillation commonly uses softened softmax labels from the teacher in training the student network~\cite{Hinton, LWF2}.
However, we argue that this formulation is not suitable for Re-ID.
In classification problems, the absolute position of the samples in the feature space must be preserved to remain compatible with the learned classifiers.
On the contrary, in Re-ID, we are interested only in preserving the relative distance between samples.
Therefore, we employ a distillation loss that acts on similarity matrices to offer the model more freedom to adjust the position of the features in the learned space.

Assuming an input tensor $\Xmat$ corresponding to a mini-batch of $n$ images from the support set $\{\xvect_i\}_{i=1}^{n}$, we use the student network $\mathcal{F}$ to compute the feature representations $\Fmat= \mathcal{F}(\Xmat)\in \mathbb{R}^{n \times c}$, where c is the dimension of the feature space.
Similarly, we compute the features with the teacher network $ \bar{\Fmat}= \bar{\mathcal{F}}(\Xmat)\in \mathbb{R}^{n \times c}$.
Then, we calculate the similarity matrices $\Smat \in \mathbb{R}^{n \times n}$ and $\bar{\Smat} \in \mathbb{R}^{n \times n}$ containing the pairwise scalar product between the current features of all images in the current batch of the support set:
\begin{equation}
    \mathbf{\Smat} =  \Fmat \Fmat^{\mathsf{T}}
    \text{, and~}
    \bar{\Smat} = \bar{\Fmat} \bar{\Fmat}^{\mathsf{T}}  .
\end{equation}
Moreover, we minimize the Frobenius norm $\Vert \cdot \Vert_F$ between the similarity matrices of the teacher and the student.
The source-guided knowledge distillation loss can thus be formulated as follows:
\begin{equation}
\mathcal{L}_{KD}(\bar{\Smat},\mathbf{\Smat}) = \left\lVert \frac{\bar{\Smat}}{\lVert\bar{\Smat}\rVert} - \frac{\mathbf{\Smat}}{\lVert\mathbf{\Smat}\rVert}\right\rVert_{F}^2.
\end{equation}

\subsection{Source-Target Distribution Alignment}
\label{sec:MMD}

To achieve successful knowledge distillation over the support set, it is crucial to ensure that the selected images from the source domain are visually similar to the previously seen target images.
To this end, we introduce an additional training loss that explicitly aligns the source and the target feature distribution.
We use the Maximum Mean Discrepancy (MMD) loss \cite{MMDLoss} to reduce the domain shift by minimizing the discrepancy between the source and target domains.
Formally, given an input batch of images $\{\xvect_i^s\}_{i=1}^{n}, \{\xvect_j^t\}_{j=1}^{n}$ coming from both $\mathcal{S}$ and $\mathcal{T}^k$, we compute the feature representations from both the teacher and the student models:
$\mathbf{\bar{B}} = (\mathbf{\bar{b}}_i)_{i=1}^n , \mathbf{B} = (\mathbf{b}_j)_{j=1}^n \in \mathbb{R}^{n \times c}$,
where:
\begin{equation}
    \mathbf{\bar{b}}_i = \mathcal{\bar{F}}(\xvect_i^s), \text{and~}\mathbf{b}_j=\mathcal{F}(\xvect_j^t).
\end{equation}
As shown in~\cite{MMDLoss}, assuming a positive semi-definite kernel $K$, the MMD loss can be empirically estimated as follows:
\begin{align}
    \nonumber
    \mathcal{L}_{MMD}(\mathbf{\bar{B}},\mathbf{B}) =\frac{1}{n^2}\sum_{i,j=1}^{n} [ K(\mathbf{\bar{b}}_i, & \mathbf{\bar{b}}_j) + K(\mathbf{b}_i,\mathbf{b}_j) \\
    \label{eqn:MMD_loss}
    & - 2\text{K}(\mathbf{\bar{b}}_i,\mathbf{b}_j)].
\end{align}
We follow the common practice and employ the Gaussian kernel \cite{MMDkernel} with bandwidth parameter $\sigma$:
\begin{equation}
\label{eqn:kernel_trick}
K(\mathbf{\bar{b}}_i,\mathbf{b}_j) =\exp \left(-\frac{\left\lVert\mathbf{\bar{b}}_i-\mathbf{b}_j\right\rVert^2}{2\sigma^2} \right),
\end{equation}
where we set the bandwidth $\sigma$ to the estimated variance of each minibatch as in \cite{MMDkernel}.

\subsection{Incorporating Pseudo-Labeling into \method. }
\label{sec:IoM}

We now detail how we integrate three state-of-the-art pseudo-labeling-based frameworks into \method: MMT~\cite{MMT}, SpCL~\cite{SpCL} and IDM~\cite{IDM}.

\noindent\textbf{MMT} employs two networks $\mathcal{F}_1$ and $\mathcal{F}_2$ instead of a single feature extractor $\mathcal{F}$ as discussed above.
The classifier $C_1$ for the feature extractor $\mathcal{F}_1$ is trained to predict the clustering labels obtained from $\mathcal{F}_2$ and vice-versa.
Mean teacher networks  $\bar{\mathcal{F}}_1$ and $\bar{\mathcal{F}}_2$ are introduced.
In addition to the cross-entropy loss $\mathcal{L}_{ce}$, and the triplet loss $\mathcal{L}_{tri}$ introduced in the \emph{strong baseline} \cite{SB}, the two networks $\mathcal{F}_1$ and $\mathcal{F}_2$ are also optimized using a soft classification loss $\mathcal{L}_{sce}$ and a soft triplet loss $\mathcal{L}_{stri}$ with their mean networks \cite{LWF2}.
Finally, $\mathcal{L}_{ReID}$ is a weighted sum of the four aforementioned losses.
To integrate MMT into \method, the two similarity matrices $\mathbf{S}_1$ and $\mathbf{S}_2$ are estimated using respectively $\mathcal{F}_1$ and $\mathcal{F}_2$ as student networks from a support set mini-batch.
Similarly, two teacher similarity matrices $\bar{\Smat}_1$ and $\bar{\Smat}_2$ are estimated from the two mean teachers.
The total knowledge-distillation loss is defined as the sum of $\mathcal{L}_{KD}(\bar{\Smat}_1,\mathbf{\Smat}_1)$ and $\mathcal{L}_{KD}(\bar{\Smat}_2,\mathbf{\Smat}_2)$.
In the same way, $\mathcal{L}_{MMD}$ is jointly optimized on the source and the target domains in the feature spaces of both student-teacher couples $(\mathcal{F}_1,\bar{\mathcal{F}_1})$ and $(\mathcal{F}_2,\bar{\mathcal{F}_2})$.

\noindent\textbf{SpCL} adopts a contrastive training scheme in the feature space over a hybrid memory that is continually updated by the estimated pseudo-labels. 
The hybrid memory stores three types of feature representations: 1) the centroids for every class of the source domain, 2) the centroids for every cluster from the target domain, and 3) the feature representations of the outliers.
Finally, $\mathcal{L}_{ReID}$ is a contrastive loss that jointly distinguishes classes, clusters, and unclustered instances in the feature space of the hybrid memory.
For more details, the readers are referred to \cite{SpCL}.
The integration of SpCL into our \method~ is straightforward. We first add the teacher model, which is the EMA of the fine-tuned model. Then, for each new task, the support set is constructed to add $\mathcal{L}_{KD}$ and $\mathcal{L}_{MMD}$ to the \method~pipeline.

\noindent\textbf{IDM} is based on a module designed to generate intermediate domain representations by mixing the hidden representations of the source and target domains.
Network training is regularized with additional losses, which promote diversity among the domain variables and ensure that the intermediate domain lies between the source and target domains.
To integrate IDM into our \method~framework, we first add a teacher model which is obtained through EMA over the model's weights, including the IDM module.
Then, during the optimization, we sum the two losses of \method, $\mathcal{L}_{KD}$ and $\mathcal{L}_{MMD}$, to the IDM losses.




\section{Experiments and Results}
\label{sec:expe}

This section introduces the datasets used in the current work, the evaluation protocol, the implementation details, as well as the results and discussions of \method.
We compare our algorithm against four state-of-the-art approaches for UDA
for person Re-ID: the \textit{strong baseline} \cite{SB}, MMT \cite{MMT}, SpCL \cite{SpCL}, and IDM \cite{IDM}.
Finally, we perform a set of ablation studies to analyze each component of \method, including the construction of the support set, the choice of the teacher, and the loss functions.
In particular, we compare our KD loss  $\mathcal{L}_{KD}$ with alternatives \cite{SP-KD,AT} previously introduced in the literature for similar tasks.


\begin{table*}[ht]
\centering
\resizebox{0.99\textwidth}{!}{
\setlength{\tabcolsep}{8pt}
\begin{tabular}{lcccccccc}
\toprule
\multirow{2}{*}{\textbf{Method}}  & \multicolumn{2}{c}{\textbf{MS $\to$ M}} & \multicolumn{2}{c}{\textbf{MS $\to$ C}} & \multicolumn{2}{c}{\textbf{M $\to$ MS}}& \multicolumn{2}{c}{\textbf{RP $\to$ M}} \\ 
& \multicolumn{1}{c}{\textbf{mAP}} & \multicolumn{1}{c}{\textbf{Rank-1}} & \multicolumn{1}{c}{\textbf{mAP}} & \multicolumn{1}{c}{\textbf{Rank-1}} & \multicolumn{1}{c}{\textbf{mAP}} & \multicolumn{1}{c}{\textbf{Rank-1}} & \multicolumn{1}{c}{\textbf{mAP}} & \multicolumn{1}{c}{\textbf{Rank-1}}   \\
\midrule
Strong Baseline \cite{SB}  &  $51.4_{\pm \num{1.8}}$& $72.3_{\pm \num{0.5}}$ &   $5.3_{\pm \num{1.2}}$ & $4.3_{\pm \num{1.9}}$ & $6.1_{\pm \num{0.1}}$ & $18.1_{\pm \num{0.3}}$ & $43.1_{\pm \num{1.3}}$               &      $67.6_{\pm \num{1.6}}$       \\
MMT \cite{MMT} &  $65.8_{\pm \num{0.1}}$ &    $83.7_{\pm \num{0.1}}$      &        $32.2_{\pm \num{1.6}}$ &    $32.2_{\pm \num{2.4}}$   & $15.1_{\pm \num{1.9}}$  &    $36.9_{\pm \num{0.1}}$ &   $58.7_{\pm \num{0.7}}$             &    $77.5_{\pm \num{0.1}}$ \\
SpCL \cite{SpCL} & $53.5_{\pm \num{0.4}}$ &   $76.0_{\pm \num{0.3}}$ &  $15.6_{\pm \num{3.1}}$ &  $15.7_{\pm \num{1.7}}$ & $14.7_{\pm \num{0.2}}$  & $36.7_{\pm \num{2.3}}$ &   $50.5_{\pm \num{2.8}}$              &     $72.1_{\pm \num{3.5}}$           \\
IDM \cite{IDM} & ${57.5}_{\pm \num{0.2}}$ &   ${78.6}_{\pm \num{0.2}}$ &  ${8.3}_{\pm \num{0.2}}$ &  ${7}_{\pm \num{0.3}}$ & ${7.9}_{\pm \num{0.5}}$  & ${21.5}_{\pm \num{0.1}}$ & $60.8_{\pm \num{0.2}}$ &     $80.4_{\pm \num{0.1}}$        \\ \hdashline
S2P-MMT (\textcolor{red}{ours}) &$\underline{70}_{\pm \num{0.4}}$ & $\underline{87.1}_{\pm \num{0.4}}$ &  $\textbf{40.4}_{\pm \num{0.8}}$ & $\textbf{42.4}_{\pm \num{0.9}}$ &$\underline{19.5}_{\pm \num{0.1}}$ &   $\underline{43.3}_{\pm \num{0.7}}$ & $\underline{61.4}_{\pm \num{0.1}}$             &  $\underline{81}_{\pm \num{0.2}}$ \\
S2P-SpCL (\textcolor{red}{ours}) &  ${69.1}_{\pm \num{0.1}}$  &   ${87.1}_{\pm \num{0.1}}$  & ${\underline{34.3}}_{\pm \num{0.3}}$ &    ${\underline{35.1}}_{\pm \num{0.5}}$ & $\textbf{20.2}_{\pm \num{0.1}}$ & $\textbf{46.1}_{\pm \num{0.2}}$ & ${59}_{\pm \num{0.1}}$            &    ${80.5}_{\pm \num{0.2}}$ \\
S2P-IDM (\textcolor{red}{ours}) &  $\textbf{71.3}_{\pm \num{0.1}}$ &   $\textbf{88.0}_{\pm \num{0.1}}$ &  ${17.5}_{\pm \num{0.5}}$ &  ${16.6}_{\pm \num{0.5}}$ & ${14.2}_{\pm \num{0.3}}$  & ${33.9}_{\pm \num{0.2}}$ & $\textbf{70.2}_{\pm \num{0.2}}$ &     $\textbf{86.1}_{\pm \num{0.4}}$  \\
\bottomrule
\end{tabular}
}
\caption{Performance of S2P~and four state-of-the-art methods in the last task in three real-to-real and one synthetic-to-real OUDA-Rid tasks. The best and second-best methods on each dataset are highlighted in {\textbf{bold}} and \underline{underlined}, respectively.}
\label{tbl:real-to-real}
\end{table*}

\noindent \textbf{Datasets.} 
We evaluate \method~on four widely used person Re-ID datasets in domain adaptation:
\begin{itemize}[leftmargin=*]
    \item \textit{Market 1501} (M) \cite{Market} has $1,501$ identities captured by six cameras. It includes $32,668$ images, with $12,936$ training images from $751$ identities and $19,732$ test images from the remaining $750$ identities. The official protocol matches $3,368$ query images to the test images.
    \item \textit{MSMT17} (MS) \cite{PTGAN} includes videos from $15$ cameras. The training set has $32,621$ images of $1,042$ identities, while the test set comprises $11,659$ query images and $82,161$ gallery images from $3,060$ identities.
    \item \textit{CUHK03} (C) \cite{cuhk03} comprises $14,097$ photos of $1,467$ individual identities from six cameras, each identity is recorded by two cameras. It includes both manual and automatic bounding boxes. We utilize manually-annotated bounding boxes for training and testing.
    \item \textit{RandPerson} (RP) \cite{RandPerson} is a synthetic dataset containing $8,000$ identities and $1,801,816$ images.
    We use a subset of $132,145$ images from the original $8,000$ identities.
\end{itemize}



\noindent\textbf{Evaluation protocol.}
We follow the experimental protocol introduced in~\cite{cvprw}.
We evaluate the performance of all methods using the standard training/testing splits proposed by the original authors for \textit{Market 1501} and \textit{MSMT17}.
In \textit{CUHK03}, we use a more challenging testing protocol proposed in \cite{DBLP}, which consists of splitting the dataset into $767$ and $700$ identities for training and testing, respectively.
RP is always used as a source dataset in this work.

We evaluate \method~for OUDA-Rid in several real-to-real and synthetic-to-real configurations: MS$\to$M, MS$\to$C, M$\to$MS, and RP$\to$M.
These configurations are widely used in the literature \cite{MMT,SpCL,RandPerson} and illustrate domain shifts of diverse difficulties.
For each configuration, we randomly and uniformly split the training identities into five subsets, corresponding to five tasks for OUDA-Rid, each having a distinct set of identities.
We also perform additional experiments where we increase the number of tasks in the target domain, which are detailed in the supplementary material due to space limitations.




We adopt the commonly used metrics for evaluation in Re-ID \cite{MMT, SpCL}: mean Average Precision (mAP) and CMC Rank-1 \cite{Market} accuracies.
These metrics are computed on the entire test set of the target domain after each task during the online adaptation process.
We report the average mAP and Rank-1 over three repetitions with different seeds.




\noindent\textbf{Implementation details. }
We follow the common practices in the UDA person Re-ID field by adopting ResNet50 \cite{resnet} pre-trained on ImageNet \cite{imagenet} as a backbone.
We employ the features computed after the global average  pooling layer.
We use DBSCAN for clustering, which is commonly employed in pseudo-labeling methods because it requires no prior assumption on the number of clusters.
For each new task, Adam \cite{adam} optimizer is adopted with an initial learning rate (LR) equal to $3.5e\!-\!4$, a linear LR scheduler, and weight decay of $5e\!-\!4$ \cite{MMT, SpCL}.
Same as \cite{cvprw}, the number of epochs per task is set to $20$.
For the EMA, we follow \cite{MMT} and set $\alpha$ to $0.999$ to update the teacher model parameters.
Finally, all the images are resized to $256 \times 128$ before being fed into the backbone (or backbones for MMT), and the batch size was set to $64$ corresponding to $16$ different identities with $4$ images per ID.

\subsection{Quantitative Results}

\begin{figure*}
    \centering
    \includegraphics[width=\textwidth]{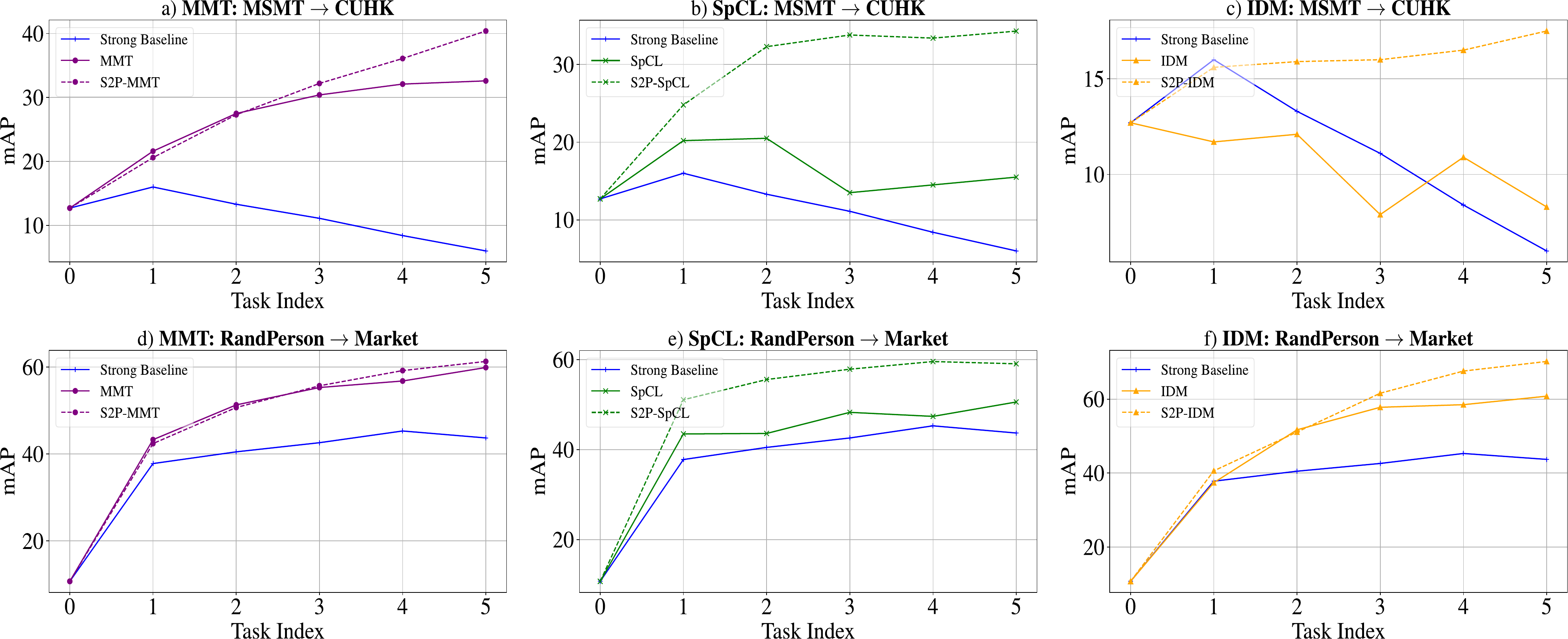}
    \caption{Comparison of \method~with four state-of-the-art methods in terms of mAP vs. task index in two different OUDA-Rid tasks, MSMT$\to$CUHK and RandPerson$\to$Market.}
    \label{fig:results}
\end{figure*}

\noindent\textbf{Comparison with the state of the art.}
Table \ref{tbl:real-to-real} reports the mAP accuracy and CMC Rank-1 score obtained at the end of training with all methods in three \textit{real-to-real} configurations: MS$\to$M, MS$\to$C, M$\to$MS, and one \textit{synthetic-to-real} RP$\to$M.
The reported metrics are computed at the end of the adaptation process in each case.
The low scores of the \emph{strong baseline} are due to the presence of the domain shift, which cannot be appropriately addressed with this method.
The state-of-the-art UDA methods MMT, SpCL and IDM struggle when deployed in the OUDA-Rid setting.
The drop in performances of MMT, SpCL and IDM is partially explained by the presence of catastrophic forgetting.
Furthermore, MMT outperforms SpCL and IDM in almost all configurations, showing that their student-teacher framework is well suited to OUDA-Rid.

Table \ref{tbl:real-to-real} shows that \method-MMT, \method-SpCL or \method-IDM outperforms all previous state-of-the-art UDA methods in OUDA-Rid over all configurations.
For example, \method~improves the mAP of SpCL from $15.6$ to $34.3$ and from $14.7$ to $20.2$ in MS$\to$C and M$\to$MS, respectively.

As for IDM, our \method~significantly improves its performances, from $8.3$ to $17.5$ and from $7.9$ to $14.2$, in the same configurations: MS$\to$C and M$\to$MS.
Finally, for MMT, \method~improves the mAP, from $32.2$ to $40.4$ and from $15.1$ to $19.5$, in MS$\to$C and M$\to$MS, respectively.
The gain for SpCL and IDM is greater than for MMT because MMT already integrates a teacher in its knowledge distillation loss function (soft cross entropy and soft triplet loss), whereas SpCL and IDM are only optimized using hard pseudo labels without any refinement.

Similarly, we can see that in the \textit{synthetic-to-real} scenario RP$\to$M, \method~noticeably improves the  performance of the three state-of-the-art methods. 
\method~improves: from $58.7$ to $61.4$, from $50.5$ to $59$, and from $60.8$ to $70.2$ the performances of MMT, SpCL, and IDM, respectively.
These results demonstrate that \method~can be successfully deployed in OUDA-Rid applications where we cannot have access to a real and well-annotated dataset for the source domain\footnote{Additional experiments in different configurations can be found in the supplementary materials.}.


\noindent\textbf{Continual behavior.}
To delve deeper into the analysis on the continual behavior of the different methods, we compare in Fig. \ref{fig:results} the mAP at the end of each task before and after incorporating the three state-of-the-art methods MMT, SpCL and IDM into our \method~framework.
For this analysis, we choose two different configurations: MS$\to$C (Fig. \ref{fig:results}-a, -b, and -c) and RP$\to$M (Fig. \ref{fig:results}-d, -e and -f).
In general, the low performances of the direct inference (\ie~the mAP at task $0$) and the \emph{strong baseline} show that the chosen configurations are of varying degrees of difficulty. 
Fig. \ref{fig:results} also shows the effect of catastrophic forgetting as a drop in performance in new tasks in several situations.
For example, the \textit{strong baseline} presents degradation of performance in both configurations in new tasks.
Similarly, SpCL and IDM both lose accuracy when confronted with new incoming data due to catastrophic forgetting and domain shift in the later tasks.
For MS$\to$C configuration: in b) the mAP of SpCL goes from 20.5 in the second task to 13.5 in the third task, while in c) the performance of IDM drops from $10.9$ to $8.3$ in the last task. Finally, for MMT we can notice in a) that the performance reaches an undesirable plateau after the third task in the same configuration.
This shows that the knowledge acquired during the first stages of OUDA-Rid is lost during the adaptation process. 
Furthermore, the fluctuations of the mAP of SpCL and IDM in b), c), e) and f) in Fig. \ref{fig:results} illustrate the inability of the models to maintain a general structure of the feature space that captures the whole target domain distribution.

On the contrary, \method-MMT, \method-SpCL and \method-IDM show a steady improvement in performance on the two configurations.
Specifically, all the three methods achieve better performance when learning later tasks when incorporated into our \method~ framework and deliver consistent results across the different configurations.

\begin{table*}[]
\centering
\resizebox{\textwidth}{!}{
    \begin{tabular}{cc|cccccccc|cccccccc}
    \toprule 
    \multirow{2}{*}{} & \multirow{2}{*}{}  &  \multicolumn{8}{c|}{\textbf{\method-SpCL}}   & \multicolumn{8}{c}{\textbf{\method-MMT}} \\ 
     \multirow{2}{*}{ $\mathcal{L}_{MMD}$} & \multirow{2}{*}{$\mathcal{L}_{KD}$}                        &  \multicolumn{2}{c}{\textbf{MS $\to$ M}} & \multicolumn{2}{c}{\textbf{MS $\to$ C}} &  \multicolumn{2}{c}{\textbf{M $\to$ MS}} &  \multicolumn{2}{c|}{\textbf{RP $\to$ M}} &  \multicolumn{2}{c}{\textbf{MS $\to$ M}} & \multicolumn{2}{c}{\textbf{MS $\to$ C}} &  \multicolumn{2}{c}{\textbf{M $\to$ MS}} &  \multicolumn{2}{c}{\textbf{RP $\to$ M}} \\ 
    & & \multicolumn{1}{c}{\textbf{mAP}} & \multicolumn{1}{c}{\textbf{Rank-1}} & \textbf{mAP} & \multicolumn{1}{c}{\textbf{Rank-1}} & \multicolumn{1}{c}{\textbf{mAP}} & \multicolumn{1}{c}{\textbf{Rank-1}} & \multicolumn{1}{c}{\textbf{mAP}} & \multicolumn{1}{c|}{\textbf{Rank-1}} & \textbf{mAP} & \multicolumn{1}{c}{\textbf{Rank-1}} & \textbf{mAP} & \multicolumn{1}{c}{\textbf{Rank-1}} & \multicolumn{1}{c}{\textbf{mAP}} & \multicolumn{1}{c}{\textbf{Rank-1}} & \multicolumn{1}{c}{\textbf{mAP}} & \multicolumn{1}{c}{\textbf{Rank-1}} \\ 
    \midrule
    \ding{55}  & \multicolumn{1}{c|}{\ding{55}}    & $53.5$ &   $76.0$ &  $15.6$ &  $15.7$ & $14.7$  & $36.7$ &   $50.5$              &     $72.1$  & $65.8$ &    $83.7$      &        $32.2$ &    $32.2$   & $15.1$  &    $36.9$ &   $58.7$             &    $77.5$ \\
    {\ding{51}}  & \multicolumn{1}{c|}{{\ding{55}}}  & \multicolumn{1}{c}{$62.4$} & $82.9$  &  $24.1$ &  $23.6$ &  $15.2$ &  $38.5$ & \multicolumn{1}{c}{$55.4$} & $77.5$ & \multicolumn{1}{c}{$62.6$} & $81.4$  &  $27.4$ &  $26.4$  & $15.3$ & \multicolumn{1}{c}{$37$}  &  $60.8$  & \multicolumn{1}{c}{$80.2$} \\
    {\ding{55}}    & \multicolumn{1}{c|}{{\ding{51}}}     & \multicolumn{1}{c}{$65.1$}  & ${85.1}$ &   $28.2$ & $26.7$ & $16$ & $40$ & \multicolumn{1}{c}{$55.5$}  & $78.9$ & \multicolumn{1}{c}{$67$}  & $85.5$ &   $35.2$ & $35.1$  & $17.8$ & \multicolumn{1}{c}{$41.1$}  &  $60.4$  & \multicolumn{1}{c}{$80.1$} \\
    {\ding{51}}  & \multicolumn{1}{c|}{{\ding{51}}}   & \multicolumn{1}{c}{$\textbf{69.1}$}  & $\textbf{87.1}$  & $\textbf{34.3}$ & $\textbf{35.1}$ & $\mathbf{20.2}$ & $\mathbf{46.1}$ & \multicolumn{1}{c}{$\textbf{59}$}  & $\mathbf{80.5}$  & \multicolumn{1}{c}{$\textbf{70}$}  & $\mathbf{87.1}$  & $\mathbf{40.4}$ & $\mathbf{42.4}$  & $\textbf{19.5}$ & \multicolumn{1}{c} {$\mathbf{43.3}$} &  $\textbf{61.4}$  & \multicolumn{1}{c} {$\mathbf{81}$} \\ \bottomrule  
    \end{tabular}}
    \caption{Ablation study on the effectiveness of the $\mathcal{L}_{MMD}$ and $\mathcal{L}_{KD}$ loss functions using \method-SpCL and \method-MMT.}
\label{tbl:abl-1}
\end{table*}

Moreover, it is clear from the learning curves across all the different tasks that  \method~successfully adapts UDA methods to the continual setting OUDA-Rid, resulting in a superior learning process evolution and a solid accumulation of prior knowledge. 
\subsection{Ablation Studies}

We perform three ablation studies about: 1) the loss functions, 2) the knowledge distillation design, and 3) the choice of the teacher model.
We run those experiments with \method-SpCL as the pseudo-labeling method in OUDA-Rid configurations, namely, MS$\to$C and RP$\to$M.

\noindent \textbf{The impact of the two main losses of \method.}
The two main loss functions (KD and MMD) of \method~were introduced in Sec. \ref{sec:kd} and \ref{sec:MMD}.
In this ablation, we study the influence of different configurations of the losses $\mathcal{L}_{MMD}$ and $\mathcal{L}_{KD}$ in the performance of \method~as shown in Table \ref{tbl:abl-1}.
The performance of the baseline significantly improves in almost all the configurations by only integrating either the $\mathcal{L}_{MMD}$ or $\mathcal{L}_{KD}$.
For example, the configuration MS$\to$ M shows a gain in performance. The mAP goes from 53.5 to 62.4 with $\mathcal{L}_{MMD}$ and from 53.5 to 65.1 with $\mathcal{L}_{KD}$ for \method-SpCL.
Furthermore, combining both losses leads to an additional overall improvement in performance in all cases.

\noindent \textbf{Knowledge Distillation Design.}
We delve into our knowledge distillation mechanism focusing on two key factors: the loss function and the selection of the support set.

Regarding the support set construction, our similarity-based selection relies on a cosine similarity function $\mathcal{\xi}$ given in Eq. \eqref{eq:support-set}. We explore two different approaches to compute the support set as shown in Table \ref{tbl:abl-2}.
The first strategy employs all the images of the source domain $\mathcal{S}$ to construct the support set.
The second (Rank-1 NN) selects only the most similar image from the source domain to each previously seen image, without considering its identity's other images.
The similarity-based selection strategy $\mathcal{\xi}$ shows the best results in almost all cases as shown in Table \ref{tbl:abl-2}.
Furthermore, we compare our $\mathcal{L}_{KD}$ with two different losses that are widely used in the literature: $\mathcal{L}_{SP}$ \cite{SP-KD} which uses pairwise activation similarities to supervise the training of the student model, and $\mathcal{L}_{AT}$ \cite{AT} where only the activations are used to compute a mean squared error between the student and the teacher models.
The results of Table \ref{tbl:abl-2} allow us to draw the conclusion that our knowledge distillation design better suits the setting of OUDA-Rid and outperforms both the other knowledge distillation losses and support set selection strategies.

\begin{figure}
 \def\myim#1{ \includegraphics[width=7.0mm,height=14.0mm]{#1}}
     \centering
   \setlength\tabcolsep{0.5 pt}
   \renewcommand{\arraystretch}{0.2}
     \begin{tabular}{cccccccccccc}
 & \multicolumn{4}{c}{\footnotesize \textbf{Source (MS) $\to$ Target (C)}} & ~ & ~ & \multicolumn{4}{c}{\footnotesize \textbf{Source (RP) $\to$ Target (M)}}
\\
$\xvect^t$ &
\makecell[l]{\myim{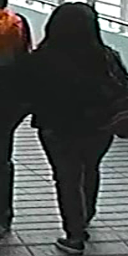}} &
\makecell[l]{\myim{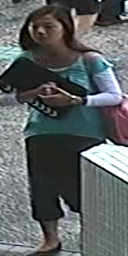}} &
\makecell[l]{\myim{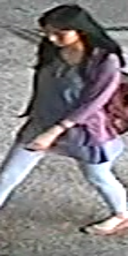}} &
\makecell[l]{\myim{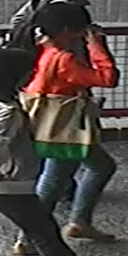}} & ~ & ~ &
\makecell[l]{\myim{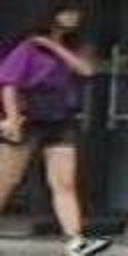}} &
\makecell[l]{\myim{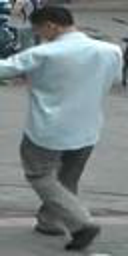}} &
\makecell[l]{\myim{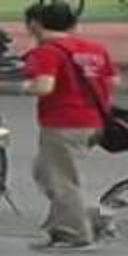}} &
\makecell[l]{\myim{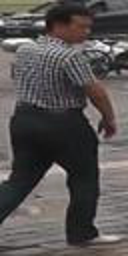}}
\\
$\mathcal{\xi}_x(\xvect^t)$&
\makecell[l]{\myim{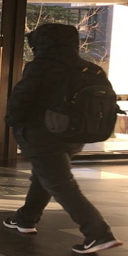}} &
\makecell[l]{\myim{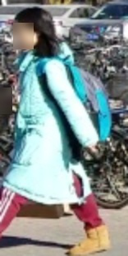}} &
\makecell[l]{\myim{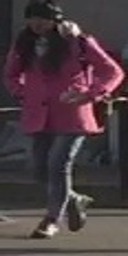}} &
\makecell[l]{\myim{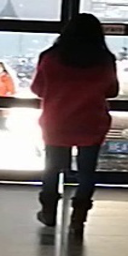}} & ~ & ~ &
\makecell[l]{\myim{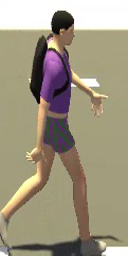}} &
\makecell[l]{\myim{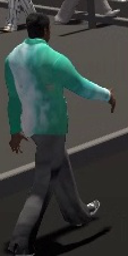}} &
\makecell[l]{\myim{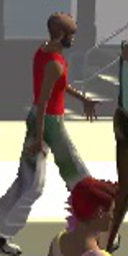}} &
\makecell[l]{\myim{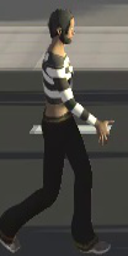}} &
      \end{tabular}
\caption{The support set construction based on the similarities between the source domain MS (RP respectively) and the target domain C (M respectively).}
\label{fig:support_s_c}
\end{figure}

To qualitatively illustrate the construction of our support set, in Fig. \ref{fig:support_s_c}, we show some random samples of the support set for MS$\to$C and RP$\to$M, where $\xvect^t$ is the image in the target domain and $\xi_x(\xvect^t)$ is its most similar image in the source domain.

\begin{table}[]
\centering
\resizebox{0.99\columnwidth}{!}{
\begin{tabular}{cccccc}
\toprule
 \textbf{Dist.} & \multirow{2}{*}{\textbf{Support Set}}      &  \multicolumn{2}{c}{\textbf{MS $\to$ C}} &  \multicolumn{2}{c}{\textbf{RP $\to$ M}}                        \\ 
  \textbf{Loss}&& \multicolumn{1}{c}{\textbf{mAP}} & \multicolumn{1}{c}{\textbf{Rank-1}} &  \multicolumn{1}{c}{\textbf{mAP}} & \multicolumn{1}{c}{\textbf{Rank-1}}  \\ \midrule
$\mathcal{L}_{KD}$&Source Domain $\mathcal{S}$   & $29.3$ & $28.1$ & $56.3$ & $78.3$      \\
$\mathcal{L}_{KD}$&Rank-1 NN   & \multicolumn{1}{c}{$29.8$} & $29.6$ & $56.4$ & $78.9$     \\
$\mathcal{L}_{KD}$&Similarity-based $\xi$   & \multicolumn{1}{c}{$\textbf{34.3}$}  & $\textbf{35.1}$  & $\textbf{59}$ & $\textbf{80.5}$\\
\midrule
$\mathcal{L}_{SP}$ \cite{SP-KD} & Similarity-based $\xi$   & \multicolumn{1}{c}{$26.5$}  & $25$  & $55.4$ & $78.8$
\\
$\mathcal{L}_{AT}$ \cite{AT} & Similarity-based $\xi$   & \multicolumn{1}{c}{$26.4$}  & $25.6$  & $55.5$ & $78.8$
\\ \bottomrule   
\end{tabular}
}
\caption{{Ablation study on the design of our knowledge distillation mechanism using \method-SpCL. We assess the impact of two key factors: the loss function and the selection function of the support set. See text for details. 
}}
\label{tbl:abl-2}
\end{table}

\noindent \textbf{The choice of the teacher.}
As described in Sec.~\ref{sec:kd} for \method, knowledge distillation is performed with a teacher network obtained via EMA updates.
In this ablation study, we investigate alternative solutions for the choice of the teacher model as shown in Table \ref{tbl:abl-3}.
We analyze three teacher models: 1) at the start of each task $t$, the teacher is frozen and initialized by the weights of the fine-tuned model on the previous task $\mathcal{F}_{t-1}$; 2) the teacher is an EMA of the student model, being updated only at the end of the previously seen tasks $\mathcal{\bar{F}}_{t-1}$; and 3) the mean teacher $\bar{\mathcal{F}}$ obtained via EMA after each iteration (\ie, one mini-batch pass) as in Sec~\ref{sec:kd}.
The results in Table \ref{tbl:abl-3} suggest that the choice of the teacher model is highly critical to alleviating the problem of catastrophic forgetting and that the proposed solution outperforms other alternatives.

\begin{table}[]
\centering
\resizebox{0.99\columnwidth}{!}{
\begin{threeparttable}
\begin{tabular}{ccccc}

\toprule
   \multirow{2}{*}{\textbf{Teacher Model}}                                                     &  \multicolumn{2}{c}{\textbf{MS $\to$ C}}  & \multicolumn{2}{c}{\textbf{RP $\to$ M}}                       \\ 
 & \multicolumn{1}{c}{\textbf{mAP}} & \multicolumn{1}{c}{\textbf{Rank-1}} & \multicolumn{1}{c}{\textbf{mAP}} & \multicolumn{1}{c}{\textbf{Rank-1}}   \\ \midrule
Task-specific $\mathcal{F}_{t-1}$   & \multicolumn{1}{c}{$14.3$} & $14.9$ & \multicolumn{1}{c}{$28.7$}   & $57$  \\
EMA of task-specific $\mathcal{\bar{F}}_{t-1}$   & \multicolumn{1}{c}{$14.8$} & $15.1$ & \multicolumn{1}{c}{$28.3$}    &  $55.7$  \\
EMA of the student $\bar{\mathcal{F}}$   & \multicolumn{1}{c}{$\textbf{34.3}$}  & $\mathbf{35.1}$  & $\mathbf{59}$ & $\mathbf{80.5}$   \\ \bottomrule   
\end{tabular}
\end{threeparttable}}
\caption{{Ablation study on the choice of the teacher model for Knowledge Distillation using \method-SpCL.}}
\label{tbl:abl-3}
\end{table}

\section{Conclusions}
\label{sec:conclusions}

In this paper, we introduced a new Source-guided Similarity Preservation (\method) algorithm for the problem of Online Unsupervised Domain Adaptation for person Re-identification (OUDA-Rid).
\method~jointly addresses catastrophic forgetting and domain shift with a knowledge distillation mechanism that respects data privacy regulations.
This mechanism is based on a support set composed of source images similar to previously seen identities in the target dataset.
We also introduced an explicit source-target distribution alignment and a pseudo-labeling strategy to alleviate the domain shift.
We performed extensive experiments where \method~straightforwardly incorporates existing state-of-the-art UDA methods and consistently outperformed them by significant margins.
\newpage
{\small
\bibliographystyle{ieee_fullname}
\bibliography{egbib}
}

\end{document}